# CrisisFake: Benchmark Validity of AI-Generated Text Detection for Disaster Social Sensing

**Xiaoshan Zhou**[1,*], **Zaifu Zhan**[2]

[1] School of Project Management, Faculty of Engineering, The University of Sydney, Sydney, NSW 2006, Australia, xiaoshan.zhou@sydney.edu.au

[2] Department of Electrical and Computer Engineering, University of Minnesota Twin Cities, Minneapolis, MN 55455, United States, zhan8023@umn.edu

## Abstract

Disaster social sensing converts public social-media posts into evidence for situational awareness and humanitarian response, but plausible large language model (LLM)-generated posts can contaminate this information stream and distort assessments of needs, damage, and resource priorities. This study empirically investigates whether text-based detectors can reliably distinguish human-authored from LLM-generated disaster posts and what textual cues underlie their judgments. We construct CrisisFake, a Qwen2.5-7B-based dataset of 12,000 texts organized into 3,000 matched semantic units from nine disasters. Each unit contains an original human post, a minimally LLM-proofread human post, a fact-preserving LLM-generated post produced using low-rank adaptation (LoRA), and an affectively reframed version of the artificial post. A separate 6,000-text corpus spanning 42 disaster events is independently constructed to support model selection and threshold calibration. We evaluate OSM-Det, Fast-DetectGPT, Binoculars, and direct LLM judges across five model families, and further examine how disaster-domain calibration and superficial linguistic cues, such as retweet markers, user mentions, hashtags, URLs, punctuation, and text length, affect detector performance. Across fourteen frozen cross-family configurations, AUROC ranges from 0.402 to 0.517, while the best prospective recall at a calibration-derived low-false-positive operating point is only 3.6%, indicating near-chance discrimination. For OSM-Det, a disaster-calibrated linear head improves AUROC to 0.817; however, a seven-feature textual classifier alone reaches AUROC 0.784 on the original-human-versus-factual-LLM contrast, and neutralizing identified surface asymmetries reduces the corresponding linear-head AUROC from 0.733 to 0.594. The calibrated head also distinguishes factual from affectively reframed LLM posts despite unchanged provenance, indicating substantial sensitivity to linguistic form rather than authorship alone. These findings provide empirical evidence that LLM-generated text detection is largely driven by linguistic cues and remains insufficiently robust for short-form disaster social media. A reliable trust gate should incorporate platform-level regulation and multimodal corroboration to better safeguard disaster social sensing.

## 1. Introduction

Climate change is increasing the frequency and intensity of several weather and climate extremes, while growing exposure and vulnerability continue to amplify disaster losses [1–3]. During fast-moving emergencies, responders need timely, local information about affected people, damaged infrastructure, disrupted access, and unmet needs. Social media can supply such information before conventional field reports are complete, particularly when roads, telecommunications, or official reporting channels are disrupted [4–6].

Disaster social sensing emerged from efforts to convert these public messages into situational awareness. Foundational corpora and benchmarks, such as CrisisNLP, CrisisMMD, and CrisisBench, enabled models to identify informative posts and humanitarian categories across events [7–10]. Operational studies demonstrate the value of this stream. Geolocated crowdsourced messages supported needs mapping after the 2010 Haiti earthquake [11]. During Hurricane Harvey, residents posted rescue requests containing addresses and other location cues; subsequent work mapped their spatiotemporal distribution [12], and VictimFinder identified rescue-request posts with an F1 score of 0.919 [13]. More recent systems use transformer and large language model (LLM) methods to classify, geolocate, aggregate, and summarise disaster reports [14–18].

Generative artificial intelligence (AI) however makes this information channel a double-edged resource. Models can produce messages that resemble eyewitness reports [19], and synthetic posts can enter the same downstream pipelines as genuine observations. Disaster misinformation has already been associated with distorted perceptions of severity, erosion of trust, inappropriate public action, overloaded emergency services, and misallocated response resources [20–24]. Consider a pipeline that identifies and geocodes genuine Harvey-style rescue requests [11–13]. A synthetic post claiming that residents are trapped at a plausible address could become a rescue waypoint; many such posts could create an artificial cluster of apparent need and divert boats or medical teams. Fabricated reports of bridge collapses, blocked evacuation routes, or supply shortages could likewise distort accessibility maps and demand estimates.

The empirical gap is therefore consequential. LLMs can generate persuasive, localised, human-like misinformation [25–28], yet there is little systematic evidence on whether AI-text detectors can flag such content in disaster social sensing. Existing benchmarks show that detector performance varies across generators, domains, platforms, languages, text lengths, and post-editing operations [28–33]. Fine-tuned LLMs can produce social-media posts whose detectability falls sharply when the detector lacks access to the attacker's generator [28]. TweepFake established the importance of short-form social-media detection, but its generators largely predate current instruction-tuned LLMs [34]. Short crisis posts therefore combine a high-stakes application with conditions in which text-only provenance evidence is intrinsically limited.

This study addresses that gap with a paired, event-disjoint disaster dataset. Each semantic unit appears in four forms: original human text (H0), minimally LLM-proofread human text (H1), factual AI-generated text based on the same verified facts (A0), and a controlled affective transformation of the AI text (A1). The paired structure separates provenance from two transformations that leave provenance unchanged. We evaluate a released social-media detector, two zero-shot likelihood-based strategies, and direct LLM-as-judge classification across multiple model families. We then calibrate thresholds on independent disaster events, train a supervised linear readout from frozen representations, and test whether apparent gains survive artifact diagnostics and targeted neutralisation.

The paper makes four contributions. First, it introduces an event-disjoint, four-condition design that distinguishes human authorship, minimal AI assistance, AI generation, and controlled affective framing. Second, it compares fourteen frozen cross-family detector configurations and OSM-Det at native and independently calibrated operating points. Third, it shows why strong supervised benchmark performance can be misleading by comparing a frozen-encoder linear head with a surface-feature-only classifier, a provenance-focused diagnostic head, and a neutralised rerun. Fourth, it reframes trust in disaster social sensing as an evidence-assessment problem rather than a binary authorship decision. Social sensing can remain valuable as a stream of potentially actionable observations, but its trustworthiness cannot be secured by placing a text-only AI detector in front of the stream. Trust must attach to corroborated claims, accountable sources, contextual and network evidence, and, where available, stronger provenance mechanisms.

# 2. Literature Review

## 2.1 Disaster Social Sensing and Crisis Informatics

Disaster social sensing sits within the broader field of crisis informatics, which studies how affected populations, volunteers, response organisations, and computational systems collectively produce and use information during emergencies. The foundational literature established both the opportunity and the processing problem. Vieweg et al. showed that microblog posts during natural hazards contain situational details not always available through official channels [35], while Imran et al.'s widely cited survey organised crisis-message processing around information extraction, classification, clustering, and related natural language processing (NLP) tasks [36]. Reuter and Kaufhold's fifteen-year review subsequently situated social media within the institutional and volunteer practices of emergency management rather than treating it as an autonomous sensor network [37]. An interdisciplinary review of 304 studies by Zhang et al. further synthesised three recurring functions: acquiring situational awareness, supporting self-organised peer-to-peer help, and enabling authorities to hear from the public [38]. More recent reviews show the computational side of the field moving from conventional machine learning toward deep neural and transformer-based analysis [39], while a 2026 review of 120 studies explicitly characterises the emerging integration of LLMs with social-media-based disaster management [40].

The operational value is clearest when social media is connected to a response task rather than treated as a generic text-classification benchmark. In Haiti, geolocated crowdsourced messages describing needs and locations supported Search and Rescue teams and aid provision [11]. During Hurricane Harvey, social-media rescue requests revealed where residents were asking for help and allowed researchers to examine the social and geographic characteristics of communities requiring rescue [12]. VictimFinder formalised the first step of that workflow: given a stream containing many irrelevant posts, identify messages that actually request rescue so that addresses can subsequently be extracted and geocoded [13]. Similarly, distinguishing a casualty report from an infrastructure-damage report, or a request for water from a general expression of concern, changes what information is routed to responders.

Methodologically, the field has evolved through successive generations of NLP. Human-annotated resources such as CrisisNLP and harmonised benchmarks such as CrisisBench enabled supervised models to learn informativeness and humanitarian categories across multiple events [7,8]. Convolutional neural classifiers improved crisis-message classification over earlier feature-engineered approaches [9]. Transformer models then shifted the emphasis toward contextual representation: CrisisBERT introduced a crisis-specific transformer for event detection and recognition [14], and BERT-based VictimFinder models substantially improved rescue-request identification [13]. CrisisTransformers extended this domain adaptation strategy by pretraining language models and sentence encoders on more than 15 billion tokens from over 30 crisis events and evaluating them across 18 public crisis datasets [15]. This progression reflects a move from manually engineered lexical signals, through task-specific deep classifiers, to reusable contextual representations trained on crisis language itself.

The newest stage uses generative and instruction-tuned LLMs to classify posts and structure them into decision-oriented information. Cantini et al. used prompt-based LLMs to identify citizen-reported issues such as damaged buildings, broken gas pipelines, and flooding; geolocate posts containing geographic references; aggregate nearby reports; and generate information-rich summaries for emergency services and news organisations [16]. El Fekih Zguir et al. introduced an LLM-based framework that classifies requests and offers into fine-grained categories of supplies, personnel, and actions while estimating actionability [17]. CrisisSense-LLM moves further toward unified multi-label disaster classification through instruction fine-tuning [18]. Crisis analytics is evolving from "is this post relevant?" toward "what is needed, where, by whom, and how urgently?", which is the exact kind of pipeline in which corrupted input provenance can have downstream operational consequences.

This evolution also exposes two methodological requirements for the present study. First, evaluation must separate disaster events because locations, entities, hashtags, reporting styles, and humanitarian needs are strongly event-specific; random text-level splitting can therefore reward event memorisation rather than generalisation [7,15,41]. Second, social sensing should be evaluated as decision support. A model can achieve attractive aggregate accuracy yet still be unusable if it misses urgent rescue requests, fails on an unfamiliar event, or injects false observations into a resource-allocation workflow.

## 2.2 Detection of AI-Generated Text

AI-generated text detection has rapidly diversified beyond a single binary-classification paradigm. Recent surveys organise the field around supervised discriminative models, training-free or zero-shot statistical methods, watermarking and generation-time provenance signals, and hybrid or LLM-based approaches [42]. Supervised detectors learn human-versus-machine boundaries from labelled corpora. OSM-Det is particularly relevant here because it targets online social media across multiple platforms and generators [43]. Other discriminative algorithms illustrate how quickly the space has expanded. Ghostbuster combines features extracted from weaker language models through structured feature search and trains a classifier without requiring access to the target generator's token probabilities [44]. OTBDetector uses contrastive learning for both machine-text detection and generator attribution under out-of-domain, unseen-model, and human-machine-manipulation settings [45]. HERO explicitly separates human-written, machine-generated, machine-polished, and machine-translated text using length-specialist models [46]. MoSEs adds stylistic experts and conditional threshold estimation to model uncertainty and heterogeneous writing styles [47], while the recent ProSSD framework combines semantic and syntactic structure in a projected statistical space and uses a likelihood-ratio-style decision statistic for cross-domain and adversarial robustness [48].

A second family avoids detector-specific supervised training and instead exploits the statistical behaviour of language models. DetectGPT uses local probability curvature around a candidate passage [49], and Fast-DetectGPT replaces repeated perturbation with a more efficient conditional probability-curvature statistic [50]. Binoculars compares a text's perplexity under a performer model with cross-perplexity between an observer and performer [51]. DNA-GPT truncates a candidate text, regenerates continuations, and measures divergence between the observed and regenerated n-grams or probability distributions [52]. GECScore uses the tendency of human text to contain more grammar-correction opportunities than LLM text as a black-box zero-shot signal [53]. Very recent work continues to search for more stable statistical signatures; NTS, for example, measures how a text's surrogate-model probability distribution responds to controlled changes in decoding temperature [54]. These methods differ substantially in computational cost and access assumptions, but all infer provenance from regularities in the text or a proxy model.

A third approach changes the problem from post hoc inference to provenance by design. Watermarking embeds a statistical signature during generation, while signed credentials or platform-side records can preserve stronger evidence about content origin [55]. These mechanisms require cooperation from model or platform providers and may be weakened by editing or paraphrasing [56]. Direct LLM judges offer another pragmatic route: an instruction-tuned model is directly asked to infer whether a passage is human- or machine-authored. Such judgments are easy to deploy but may reproduce the stylistic heuristics and domain biases that affect human raters and supervised classifiers [57].

The social-media setting is more difficult than many conventional detector benchmarks. MultiSocial was introduced because most earlier machine-generated-text benchmarks emphasised longer English prose; it spans 22 languages, five social platforms, and seven generating LLMs, and shows that platform choice affects detector training and transfer [32]. Dawkins et al. further demonstrate that fine-tuning a generator on human social-media style can sharply reduce detectability under realistic threat assumptions [28]. TweepFake similarly established a benchmark for short synthetic tweets, but its GPT-2-, RNN-, and Markov-based generators reflect an earlier generation of text synthesis [34]. Disaster posts are often only a sentence or two long and contain hashtags, abbreviations, locations, URLs, user mentions, grammatical irregularities, and rapidly changing event vocabulary. A detector that succeeds on essays or news may therefore fail because its statistical cues are weak, reversed, or confounded in short crisis communication.

We test OSM-Det as a supervised social-media detector, Fast-DetectGPT and Binoculars as influential zero-shot probability-based strategies, and direct LLM-as-judge classification while varying the underlying model family. The wider literature informs the interpretation of these choices: newer detectors introduce regeneration, grammar-error, contrastive, stylistic, and semantic-structural signals [44–48,52–54], but they also reinforce that "AI detection" is not a single stable construct. Our aim is not to crown one detector, but to test whether representative text-only provenance signals remain operationally meaningful under short disaster posts, event shift, human–AI co-writing, and dataset-construction artifacts.

No existing benchmark directly combines modern instruction-tuned generators, matched disaster semantics, minimal human-text proofreading, controlled affective reframing, and event-disjoint evaluation. The present dataset is designed to fill that gap while making the construction process itself testable.

# 3. Methodology

## 3.1 Dataset Design and Experimental Rationale

### 3.1.1 Provenance, Veracity, and Operational Risk

Provenance and veracity are related but distinct. A human can post a false rumour, an AI system can accurately restate an official warning, and a genuine victim may use an LLM to correct grammar, translate a message, or improve accessibility. Human judgments of AI authorship are themselves unreliable, and post hoc detectors can be evaded through paraphrasing or manipulation [56,57]. If an authorship detector becomes a rejection rule, false positives may suppress legitimate machine-assisted reports; conversely, labelling a message as human provides no assurance that its claim is true [58–60].

Dataset design materially affects measured detector performance. RAID varies generators, domains, decoding strategies, text lengths, and adversarial transformations [30]; DetectRL and DetectRL-X introduce realistic editing, writing errors, multilingual conditions, and other deployment-relevant shifts [31,33]. M4 and MultiSocial likewise show that transfer varies across generators, domains, languages, and platforms [29,32]. TweepFake demonstrates the special difficulty of short synthetic social-media messages [34]. These findings motivate an event-disjoint dataset that explicitly separates provenance from editing and framing transformations.

The design must also guard against shortcut learning. Under distribution shift, features that predict labels in development data may fail after deployment [61]. Collection protocols can create label-correlated lexical or formatting artifacts [62,63], allowing a model to appear successful without learning the intended construct. Explanatory analyses of "Clever Hans" predictors show that aggregate performance alone cannot distinguish a valid decision rule from an incidental one [64]. We therefore treat artifact diagnosis and construction-neutralised reruns as core validity tests rather than optional post hoc analyses.

### 3.1.2 Matched Semantic Units and Four Conditions

Each dataset unit begins with one authentic disaster post and retains a stable pair_id across every derived version. This matched structure controls the underlying event, topic, and factual content more tightly than an unmatched collection of human and AI texts. It also supports within-unit analyses of provenance-preserving transformations, an important control because detectors can be highly sensitive to syntactic perturbations even when semantics or provenance are unchanged [65].

H0 represents the target human stream: an original crisis post preserved as written, apart from uniform masking of URLs and user mentions. Retaining spelling, punctuation, slang, hashtags, and awkwardness is essential because these features are part of authentic crisis communication. H0 anchors the dataset in the language that downstream disaster systems would actually receive [7–10,41].

H1 models limited, realistic AI assistance rather than synthetic authorship. Machine-polished text is increasingly recognised as distinct from both purely human and machine-generated writing [46]. A constrained proofreading instruction corrects only spelling, grammar, punctuation, and obvious typographical errors. It permits an unchanged return and prohibits paraphrasing, stylistic improvement, and factual alteration. H1 retains human provenance and tests whether detectors mistake minor machine assistance for machine authorship, a vulnerability also highlighted by real-world editing evaluations [31,33].

A0 represents factual AI generation under semantic control. Directly prompting a model with H0 could preserve or copy human wording, confounding provenance with lexical inheritance. We instead extract a structured fact card and generate a new short post from those verified facts. This design holds the informational target approximately constant while reducing direct wording transfer. It also makes the construction mechanism explicit so that residual surface artifacts can be audited [62–64].

A1 represents a controlled change in affective framing while preserving AI provenance and the factual body of A0. Emotional expression is operationally relevant in disaster communication and can influence diffusion and collective response on social media [66]. The condition is not intended as a LLM psychological emotion benchmark. If a provenance score changes substantially when affection blended in an AI-authored post changes, it reflects the detector is responding to textual form or a variant property of authorship embedded in emotional expression.

Together, the four conditions create two provenance classes and two paired transformation controls. The human class comprises H0 and H1; the AI class comprises A0 and A1, which is treated as positive throughout. The H0-to-H1 contrast isolates sensitivity to minimal proofreading within human-authored text, and the A0-to-A1 contrast isolates sensitivity to affective framing within AI-generated text.

### 3.1.3 Condition Definitions and Research Questions

The four dataset conditions are defined as follows using the terminology fixed in the final data manifests:

H0 — original human-authored disaster posts drawn from the source corpus.

H1 — minimally LLM-proofread human posts produced from H0 under a constrained proofreading instruction.

A0 — factual AI-generated disaster posts produced from structured fact cards derived from the corresponding H0 message.

A1 — affectively framed AI-generated posts produced by applying a controlled, provenance-preserving transformation to the corresponding A0 text.

The primary task is binary provenance classification. The matched design additionally tests discriminative validity (whether a score ranks the dataset classes), operational validity (whether a threshold transfers to unseen events at tolerable error rates), and construct validity (whether the score reflects provenance rather than templates, preprocessing, or surface markers). A model may perform well on the first while failing the other two.

All partitioning is performed at the disaster-event level. Posts from one event share vocabulary, locations, named entities, hashtags, and reporting conventions; random text-level splitting can therefore reward event memorisation rather than generalisation [7,15,41]. All four versions of a semantic unit remain within the same partition.

Three research questions guide the study:

RQ1. How do established text-only detectors perform on the disaster dataset at their native operating points and after independent disaster-domain calibration?

RQ2. How does substituting the underlying language-model family alter ranking and low-false-positive operating performance?

RQ3. When a domain-adapted supervised readout performs well, does it capture provenance or dataset-construction artifacts?

## 3.2 Dataset Construction

### 3.2.1 Source Corpus and Sampling

CrisisBench was selected as the source corpus because it harmonises eight human-annotated crisis social-media datasets [7]. We removed empty texts, records with missing event labels, replacement-character corruption, exact duplicates, repost fingerprints, and near duplicates sharing the same normalised first 14 content tokens. Posts with fewer than eight words were excluded to avoid extremely short messages with insufficient textual evidence.

We selected 3,000 H0 source units by round-robin sampling across 158 strata defined by disaster event, humanitarian-information category, source dataset, and message-length tercile. Each stratum contributed at most one record per pass, prioritising broad coverage while allowing large strata to continue after smaller strata were exhausted. A per-event cap of 1,002 was specified as an anti-dominance safeguard; the largest realised event count was 602, so neither the cap nor its deterministic top-up branch was activated. Tables 1 and 2 summarise the realised source and event distributions.

Table 1. Source-dataset composition of the 3,000 H0 posts.

| **Underlying source** | **H0 posts** |
|---|---|
| CrisisMMD | 1,436 |
| CrisisLexT26 | 962 |
| CrisisNLP-CF | 531 |
| CrisisNLP-Volunteers | 71 |
| Total | 3,000 |

Table 2. Event distribution of the 3,000 H0 posts.

| **Event** | **Number** | **Event** | **Number** |
|---|---|---|---|
| 2014_california_earthquake | 602 | 2013_bohol_earthquake | 324 |
| hurricane_harvey | 514 | 2013_canada_lac_mégantic_train_crash | 202 |
| hurricane_irma | 500 | 2013_italy_sardinia | 51 |
| mexico_earthquake | 422 | 2012_venezuela_refinery_explosion | 45 |
| 2013_singapore_haze | 340 | Total | 3,000 |

### 3.2.2 H0 Original Human Condition

The length terciles contained 1,054 short, 1,052 medium, and 894 long posts. Per-stratum counts ranged from 1 to 39 (mean 19). H0 texts were copied from the source corpus without semantic or stylistic modification, and provenance metadata was retained separately from the detection field.

### 3.2.3 H1 Minimally Proofread Human Condition

H1 was produced with Qwen2.5-7B-Instruct under a constrained proofreading instruction, using greedy decoding, a maximum of 160 new tokens, and a batch size of 80. The instruction allowed the model to return a text unchanged; consequently, many H1 records are byte-identical to their H0 counterparts.

The executed prompt template was:

```
You are performing MINIMAL proofreading of a social-media message.

Correct ONLY:
- spelling mistakes,
- grammatical errors,
- punctuation errors,
- obvious typographical mistakes.

Preserve as much of the original wording as possible.

DO NOT:
- paraphrase sentences,
- improve style,
- formalize the writing,
- summarize,
```

```
- expand,
- add facts,
- remove facts,
- change the author's tone,
- change emotional intensity,
- remove slang,
- expand abbreviations,
- remove emojis,
- remove hashtags,
- change named entities,
- change numbers,
- change locations,
- change casualty information.

If the original text is already acceptable, return it unchanged.
Return ONLY the proofread message.

ORIGINAL:
{text}
```

### 3.2.4 Fact Card Extraction

To construct A0 without encouraging direct paraphrase, each H0 post was converted into a structured JSON fact card. The card represented only the factual content required for generation, such as event type, location, magnitude, casualties, damage, or reported response activity.

For example, consider the following H0 message:

```
“RT @... 82 killed as 7.2-magnitude earthquake hits Philippines ...”
```

The corresponding fact card contained:

```
Event: earthquake

Location: Philippines

Magnitude: 7.2

Reported deaths: 82
```

The generator received the fact card rather than the original wording and could produce, for example:

```
“82 killed in Philippines by 7.2-magnitude earthquake.”
```

### 3.2.5 A0 Factual AI Generation

We trained a low-rank adaptation (LoRA) module on the Qwen2.5-7B-Instruct backbone to convert disaster fact cards into realistic short social-media posts. The LoRA training data were disjoint from both the dataset and the calibration/development corpus. Training used 60,576 records and 16,308 validation examples, with prompt tokens masked at −100 so that loss was computed only on target tokens. Event sets for training, validation, and testing were asserted to be pairwise disjoint. The adapter used rank r = 16, alpha = 32, dropout = 0.05, a learning rate of $1 \times 10^{-4}$, cosine scheduling, and a 0.03 warm-up ratio. The 40.37 million trainable parameters represented 0.53% of the 7.6-billion-parameter backbone.

During inference, the adapter remained as unmerged LoRA side branches, and the tokenizer was loaded from the adapter directory to preserve the training chat template. The first pass used greedy decoding. Generations that failed quality control underwent up to three strict batched retries using nucleus sampling (temperature 0.7, top-p 0.9, maximum 220 new tokens) with deterministic seeds derived from pair_id; a deterministic fact-card renderer was used only when retries remained unsuccessful.

The executed A0 prompt template was:

```
Write one concise disaster-related social-media post based ONLY on the verified facts below.
The message should sound like a natural social-media update.
Communicate the factual information clearly.
Do not add emotional commentary unless it is explicitly contained in the facts.
Do not invent casualties, locations, damage, rescue activity, numbers, causes, predictions,
or other information.
Keep the length within {min_words} to {max_words} words.

FACTS:
{fact_card}

Return ONLY the post.
```

### 3.2.6 A1 Controlled Affective Framing

A1 assigned one of four balanced affect categories (fear, sadness, anger, or hope) to the corresponding A0 text (750 units per category). The transformation preserved the factual body of A0 and added only a short affective frame. The executed prompt was:

```
Rewrite the following disaster-related social-media post so that it expresses {EMOTION}
naturally.

IMPORTANT: Preserve ALL factual information.
Do NOT add or remove facts; alter casualty counts, numbers, named locations, or
infrastructure damage; or invent causes, predictions, or rescue actions.
The emotional expression should sound natural for social media rather than theatrical or
exaggerated.
Keep the message approximately similar in length.

VERIFIED FACTS:
{fact_card}

ORIGINAL FACTUAL POST:
{A0_text}

Return ONLY the rewritten post.
```

### 3.2.7 Post-processing and Quality Assurance

Human reviewers examined the four dataset conditions and identified metadata leakage, hallucinated social tokens, factual or lexical failures, and mechanical affect templates. GPT-5.6 Sol was then used to apply constrained corrections to the flagged records, after which the corrected records were rechecked by human reviewers. Table 3 summarises the issues and permitted corrections.

Table 3. First-round quality review and constrained corrections.

| Condition | Issue identified in the first version | Constrained correction |
|---|---|---|
| H0 | Some human posts had an artificial leading metadata token such as earthquake, hurricane, flood, or serialized None. | Removed only the prepended metadata artifact; otherwise preserved the human post (spelling, grammar, slang, punctuation, hashtags, style, and awkwardness were all left alone). URLs were standardized. |
| H1 | Same metadata prefix problem, for example: “explosion Fire spreads at Venezuela refinery...” | Removed the metadata prefix but preserved the actual minimal-proofreading output. |
| A0 | Leading None/event-type labels, hallucinated hashtags or mentions, mismatched URL counts, | Cleaned metadata artifacts, removed hallucinated social tokens, matched |
| | and flagged 321 seriously poor generations, such as severe factual and lexical mismatch, pathological repetition, extremely truncated responses, or generic fallback text. | URLs to the paired source, and conservatively repaired severe failed generations that were manually flagged. |
| A1 | Mechanical tagging or appending rather than a controlled affective transformation, e.g. explicit emotion hashtags (#fear/#sadness/#anger/#hope) or a simple emotion phrase added to the end. | Rebuilt the failed cases in A1 from the corrected A0 so that the factual A0 body was preserved and only a short natural affective frame was added. |

Residual metadata leakage was prevalent in the first version (43.3% of H0, 39.8% of H1, 30.8% of A0, and 30.9% of A1) and was removed because metadata fields could reveal how a record was constructed. URLs, user mentions, and hallucinated social tokens were corrected under condition-specific rules. A1 records containing explicit emotion labels, emotion hashtags, appended stock sentences, or other obvious templates were manually flagged and rebuilt from the corrected A0 text. The four affect categories remained exactly balanced.

A subsequent audit found inconsistent URL normalisation caused by a defective regular expression. We therefore applied one uniform procedure to every condition: HTML was unescaped, URLs were replaced with <URL>, user mentions were replaced with <USER>, repeated whitespace was collapsed, and leading or trailing whitespace was removed. Hashtags, punctuation, casing, spelling, dashes, and all other textual features were preserved.

Table 4 presents records from the final corrected dataset. H1 and A0 are identical in 7.0% of units, usually when a short human message and the fact-card-conditioned generation coincide. H0 and H1 are exactly identical in 51.67% of units because the proofreading instruction permits unchanged output when no correction is needed.

Table 4. Representative example of the four dataset conditions. H1 corrects “grown up to” to “grown to” while preserving the value 319, and A1 adds a brief fear-oriented frame.

| **Pair** | **Event / category** | **A1 affect** | **Condition** | **Exact post** |
|---|---|---|---|---|
| pair_000151 | Mexico earthquake / injured or dead people | | H0 | In Mexico the number of victims of an earthquake has grown up to 319 <URL> <URL> |
| | | | H1 | in Mexico the number of victims of an earthquake has grown to 319 <URL> <URL> |
| | | | A0 | At least 319 people were killed by Mexico's earthquake, officials say <URL> <URL> |
| | | Fear | A1 | A worrying development: At least 319 people were killed by Mexico's earthquake, officials say <URL> <URL> |

## 3.3 Independent Calibration and Development Corpus

### 3.3.1 Sampling and Event-Disjoint Partitioning

A separate corpus of 1,500 semantic units (6,000 texts) was constructed from events disjoint from the evaluation dataset using the same four-condition procedure. This corpus supplied all detector-threshold calibration and supervised fitting, keeping the final dataset held out for evaluation. The split manifest recorded zero event-level overlap among all fitting, selection, calibration, and benchmark partitions.

The 1,500 H0 units were selected by proportional allocation followed by an equal-quota affect assignment. We did not reuse the dataset's equal-opportunity round-robin scheme because that procedure intentionally gives small events substantial representation; for calibration, it could allow unusual small events to influence thresholds disproportionately.

A largest-remainder allocation with a one-per-event floor first selected 3,000 candidate units across 42 training events. The pool was then reduced to 1,500 units, and affect labels were assigned by round-robin tiling followed by a seeded shuffle to obtain exactly 375 units per affect category. All 42 events remained represented, and selected counts correlated 0.991 with proportional expectation. The source datasets were CrisisLex6, the Figure Eight Disaster Response dataset, CrisisNLP-Volunteers, CrisisLexT26, CrisisNLP-CF, CrisisMMD, ISCRAM13, and SWDM13 [8,10]. The 1,500 units were partitioned by event into 900 fitting, 300 model-selection, and 300 calibration units.

### 3.3.2 Quality Assurance

After H1, A0, and A1 were generated, the calibration/development corpus underwent the same manual quality review. Four correction classes were applied. First, leaked disaster-type prefixes were removed from 87.0% of H0, 83.5% of H1, 17.2% of A0, and 16.3% of A1 records; literal "None" wrappers were removed from 62.9% of A0 and 63.2% of A1. Second, normalisation was recomputed for every row, correcting unmasked URLs in 79.0% of H0, 79.3% of H1, 84.8% of A0, and 88.6% of A1. Third, 34 semantic units with empty, generic, or flagged A0 outputs were replaced by affect-matched unused units screened against verbatim or near-verbatim copying of H0. Fourth, A1 was rebuilt for every replacement unit with the same controlled transformation.

The corrected corpus passed a hard quality gate: 6,000 texts; 1,500 unique semantic units per condition; identical pair_id sets across conditions; exactly balanced affect categories; no empty texts; no duplicate texts within any condition; and no residual metadata prefixes or serialised null artifacts. H0 and H1 were identical for 59.9% of units.

## 3.4 Detection Strategies

### 3.4.1 OSM-Det

OSM-Det is a released social-media AI-text detector [43]. The loaded configuration confirmed a Longformer sequence-classification architecture with 12 hidden layers, a hidden size of 768, two output labels, and a classification token at position zero. Post text was tokenised and truncated at 512 tokens. The AI class index was fixed at 1 by configuration, and the continuous score was the softmax probability of that class. The native decision rule was the model's argmax, equivalent to thresholding the probability at 0.5.

### 3.4.2 Fast-DetectGPT

Fast-DetectGPT [50] compares the observed token log-likelihood under a scoring model against the distribution of log-likelihoods expected under a reference model. For an input sequence of tokens $x_1, \ldots, x_T$, let $q_t(\cdot)$ denote the scoring-model distribution over the vocabulary at position $t$ and $p_t(\cdot)$ the reference-model distribution at the same position, both taken from the same tokenisation of the same padded batch. Define for each position the observed log-probability of the realised next token, and the first two moments of the scoring-model log-probability under the reference distribution:

$$\ell_t = \log q_t(x_{t+1}), \qquad \mu_t = \sum_v p_t(v) \log q_t(v), \qquad \nu_t = \sum_v p_t(v) (\log q_t(v))^2 - \mu_t^2.$$

With $m_t$ the attention mask over scored positions, the implemented statistic is

$$d_{\text{Fast}} = \frac{\sum_t m_t \, (\ell_t - \mu_t)}{\sqrt{\max(\sum_t m_t \, \nu_t, \ 10^{-12})}}.$$

The numerator accumulates the masked difference between the observed log-probability and its reference expectation, and the denominator is the square root of the masked sum of reference variances, clamped for numerical stability. The AI-oriented score is this quantity without transformation, so larger values correspond to more AI-like text under the pre-specified orientation. Padding contributes nothing because all sums are masked. Sequences are truncated at 256 tokens.

### 3.4.3 Binoculars

Binoculars [51] normalises the perplexity of a text under a performer model by the cross-perplexity between an observer and the performer. As implemented, with $\text{CE}_t$ the token-level cross-entropy of the performer at scored position $t$ and $m_t$ the corresponding mask,

$$\text{PPL} = \frac{\sum_t m_t \, \text{CE}_t}{\sum_t m_t}, \qquad \text{XPPL} = \frac{\sum_t \tilde{m}_t \left(-\sum_v p_t(v) \log q_t(v)\right)}{\sum_t \tilde{m}_t}, \qquad B = \frac{\text{PPL}}{\text{XPPL}}.$$

Here $p_t$ is the observer distribution, $q_t$ the performer distribution, and $\tilde{m}_t$ the full attention mask including special tokens, matching the executed code, which computes the cross-perplexity term over all non-padding positions rather than only the shifted scored positions. Because a lower Binoculars ratio indicates machine-generated text, the AI-oriented score is the negation, $s = -B$..

### 3.4.4 Direct LLM-as-Judge

The judge task asks an instruction-tuned model to classify authorship directly. The executed semantic prompt, held fixed across all families, is:

```
You are evaluating the authorship of a disaster-related social-media post.

Decide whether the post is primarily HUMAN-WRITTEN or AI-GENERATED.

Consider only the text itself. Do not assume that emotional language,
informal grammar, spelling errors, hashtags, emojis, or polished writing
necessarily indicate either class.

Return only one of the following labels:

HUMAN
AI
```

```
POST:
{text}
```

The prompt is wrapped with each model official chat template with a generation prompt appended, and the resulting token sequence is truncated to the final 320 tokens. Rather than parsing free text, the primary score is a forced-choice likelihood. For each candidate answer string $c$ in $\{$AI, HUMAN$\}$, the implementation appends the tokenised candidate to the prompt and accumulates the summed log-probability of the candidate tokens only:

$$L_c = \sum_{t \in c} \log P(c_t \mid \text{prompt}, c_{<t}).$$

The judge logit and the reported AI score are

$$z_{\text{judge}} = L_{\text{AI}} - L_{\text{HUMAN}}, \qquad s_{\text{AI}} = \sigma\Big(\text{clip}\big(z_{\text{judge}}, -80{,}80\big)\Big) = \frac{\exp(L_{\text{AI}})}{\exp(L_{\text{AI}}) + \exp(L_{\text{HUMAN}})},$$

with clipping applied for numerical stability. In parallel, the implementation greedily generates up to four new tokens and parses a hard HUMAN or AI label, recording a malformed flag when the generated string does not contain exactly one of the two labels. The native operating point for this strategy is the predeclared forced-choice rule $s_{\text{AI}} > 0.5$.

### 3.5 Cross-Family Language-Model Evaluation

Fast-DetectGPT and Binoculars each require a matched pair of models from one family, whereas the judge requires one instruction-tuned model. We evaluated five families: tiiuae/falcon-7b, Qwen/Qwen2.5-7B, mistralai/Mistral-7B-v0.3, meta-llama/Meta-Llama-3-8B, and google/gemma-2-9b. For each family, the base checkpoint supplied the Fast-DetectGPT reference distribution and served as the Binoculars observer; the instruction-tuned checkpoint supplied the Fast-DetectGPT scoring distribution, served as the Binoculars performer, and acted as the judge. Tokeniser compatibility within each pair was verified by asserting identical vocabulary sizes and identical tokenisation of a probe sentence, because the implementation tokenises once and supplies the same token identifiers to both models. All five pairs passed.

### 3.6 Evaluation Metrics

Disaster-response use imposes demanding operating requirements. A detector must handle short, informal, event-specific posts; generalise across generators and hazards; tolerate legitimate proofreading or translation; and maintain very low false-positive rates so that genuine calls for help are not discarded. Aggregate AUROC is insufficient if recall collapses in the low-false-positive region, and supervised adaptation is unsafe if it learns construction shortcuts rather than provenance [61–64]. We therefore report ranking, calibration, and paired-transformation metrics.

We treat the AI class is positive and the human class is negative. With true and false positives and negatives denoted $TP$, $FP$, $TN$ and $FN$:

$$\text{Accuracy} = \frac{TP + TN}{TP + TN + FP + FN}, \qquad \text{Precision} = \frac{TP}{TP + FP}, \qquad \text{Recall} = TPR = \frac{TP}{TP + FN},$$

$$\text{Specificity} = TNR = \frac{TN}{TN + FP}, \qquad FPR = \frac{FP}{FP + TN}, \qquad FNR = \frac{FN}{FN + TP},$$

$$F_1 = \frac{2\,\text{Precision} \cdot \text{Recall}}{\text{Precision} + \text{Recall}}, \qquad \text{Balanced Accuracy} = \frac{TPR + TNR}{2}.$$

AUROC is reported as threshold-independent ranking performance and is interpretable as the probability that a randomly drawn AI text receives a higher score than a randomly drawn human text, with ties handled by the trapezoidal implementation in scikit-learn. A value of 0.5 corresponds to chance ranking and 1.0 to perfect ranking. A value below 0.5 indicates that the ranking is systematically reversed relative to the pre-specified score orientation. AUPRC is the area under the precision-recall curve; because the benchmark is exactly balanced between the human and AI classes, the expected value under random ranking is approximately 0.5.

Two quantities that are frequently conflated are reported separately. The descriptive metric $TPR_{\text{ROC@5\%}FPR}$ is read from the benchmark ROC curve by linear interpolation at a false-positive rate of 0.05 and characterises score discrimination only. The prospective metric is obtained by applying the frozen calibration-derived threshold $T_{\text{FPR05}}$ to the dataset and measuring the realised test false-positive and true-positive rates.

For the paired transformation analyses we define, for semantic unit $i$ and score $s(\cdot)$,

$$\Delta_i^{\text{proof}} = s(\text{H1}_i) - s(\text{H0}_i), \qquad \Delta_i^{\text{affect}} = s(\text{A1}_i) - s(\text{A0}_i).$$

Because the score is AI-oriented, a positive value indicates that the transformation made the text appear more AI-like to that detector.

## 3.7 Disaster-Domain Threshold Calibration

Detector-score distributions vary with domain, platform, text length, generator, decoding strategy, editing, and the human reference corpus [28–33]. Disaster social-media posts differ markedly from the long, polished prose used in many detector evaluations, and short inputs provide less statistical evidence. We therefore evaluated both frozen off-the-shelf operating points and thresholds derived from the independent 6,000-text calibration corpus. This design tests whether poor performance is primarily a threshold mismatch or a failure of the underlying score to rank provenance.

### 3.7.1 Native Operating Points

A native operating point is reported only where a legitimate default rule exists for the exact configuration executed. For OSM-Det this is the pretrained argmax rule at probability 0.5. For the judge strategies it is the predeclared forced-choice rule at $s_{\text{AI}} > 0.5$. For Binoculars with the Falcon pair the published threshold applies to the raw ratio and corresponds to $s > -0.8536432310785527$ under our sign convention; this threshold is specific to that published configuration and was deliberately not transferred to any other family. For Fast-DetectGPT in this configuration, and for Binoculars outside the Falcon pair, no legitimate native threshold exists and none was constructed.

### 3.7.2 Threshold Calibration

Let a decision be positive when the score satisfies $s \geq t$. On the calibration corpus, define the weighted true- and false-positive rates $TPR(t)$ and $FPR(t)$. The balanced threshold maximises the Youden statistic,

$$J(t) = TPR(t) - FPR(t), \qquad T_{\text{BAL}} = \underset{t}{\operatorname{argmax}} J(t).$$

The controlled false-positive thresholds are

$$T_{\mathrm{FPR05}} = \arg\max_{t\,:\,FPR(t)\le 0.05} TPR(t), \qquad T_{\mathrm{FPR01}} = \arg\max_{t\,:\,FPR(t)\le 0.01} TPR(t).$$

The implementation does not sweep an arbitrary grid. It computes the weighted receiver operating characteristic with *sklearn.metrics.roc_curve*, which returns the set of thresholds at which the decision changes, and then selects among those candidate thresholds. For the constrained thresholds it restricts to candidates satisfying the false-positive constraint and takes the one with the largest true-positive rate, resolving ties by the first such candidate in the returned ordering. No interpolation between candidate thresholds is performed, so every reported threshold is attainable on the calibration data. If no candidate satisfies a constraint the threshold is reported as undefined.

Because each semantic unit contributes two human texts and two AI texts, treating the four as independent would let a unit count twice on each side. The executed calibration therefore assigns every calibration row a weight of 0.5, so that within the human class H0 and H1 each contribute half of their unit total influence and within the AI class A0 and A1 do likewise. Each semantic unit thus contributes unit mass to each provenance side. These weights enter the weighted ROC computation used for threshold selection. They are not applied to the final-test metrics, which are reported unweighted over all dataset texts. Dependence among the four conditions is instead handled in the uncertainty estimates through pair-level resampling, as described in Section 3.10.

Thresholds are derived exclusively on the calibration corpus, frozen, and then applied unchanged to the evaluation dataset. The resulting test false-positive and true-positive rates constitute the prospective operational result and are reported separately from descriptive ROC quantities.

## 3.8 OSM-Det Disaster-Calibrated Linear Head

For OSM-Det, we supplemented threshold calibration with supervised refitting of the linear readout. The encoder remained completely frozen while the head was trained on disaster-domain data. This procedure adapts the scoring function as well as the operational threshold.

### 3.8.1 Frozen Representation Extraction

The OSM-Det encoder is used as a fixed feature extractor with all parameters frozen. For every text the executed pipeline runs a forward pass with hidden-state output enabled and stores 13 hidden states, comprising the embedding output and the 12 encoder blocks, each of dimension 768. Two pooled representations are stored per layer: the classification-token vector at position zero, and the attention-mask mean over non-padding positions. Inputs are the detection field truncated at 512 tokens.

### 3.8.2 Partitioning for Fitting and Calibration

The calibration corpus was divided by event into three partitions using a greedy balancing procedure with seed 42: HEAD_TRAIN with 900 units across 23 events, HEAD_SELECTION with 300 units across 10 events, and HEAD_CALIBRATION with 300 units across 9 events. All four conditions of a unit remain together.

### 3.8.3 Linear Head and Model Selection

For a pooled representation h the head is a regularised logistic model

$$P(y = \mathrm{AI} \mid \mathrm{h}) = \sigma(\mathrm{w}^{\top}\mathrm{h} + b), \qquad \sigma(z) = \frac{1}{1 + e^{-z}},$$

fitted by *sklearn.linear_model.LogisticRegression* with an L2 penalty and the lbfgs solver, maximum 2000 iterations. The objective minimised is the penalised negative log-likelihood

$$\min_{\mathrm{w},b} C \sum_{i=1}^{n} [-y_i \log p_i - (1-y_i)\log(1-p_i)] + 1/2\,\|\mathrm{w}\|_2^2,$$

in the parameterisation used by the library, where the inverse regularisation strength $C$ multiplies the data term, so that smaller $C$ corresponds to stronger regularisation. Class weights are set to balanced when the class ratio exceeds 1.1 and are otherwise left uniform.

A full grid was screened over both pooling strategies, all 13 hidden-state indices, six inverse regularisation strengths (0.001, 0.01, 0.1, 1, 10, 100) and both standardisation settings, giving 312 configurations. Where standardisation was enabled, feature means and standard deviations were estimated on HEAD_TRAIN only and applied unchanged to the other partitions,

$$\tilde{h}_j = \frac{h_j - \mu_j}{\sigma_j}.$$

Every configuration was fitted on HEAD_TRAIN and ranked by AUROC on HEAD_SELECTION, with AUPRC and balanced accuracy as successive tie-breakers. The final test data played no part in selection. The selected configuration was hidden-state index 3 with mean pooling, $C = 0.1$ and standardisation disabled, achieving AUROC 0.819 on HEAD_SELECTION.

### 3.8.4 Threshold Calibration and Final Evaluation

Head weights were fitted on HEAD_TRAIN. Thresholds were then derived on HEAD_CALIBRATION only, and applied unchanged to the dataset. The reported continuous score is the decision function of the linear head.

## 3.9 Dataset-Artifact Diagnostics

Despite the quality controls, condition-specific processing could still introduce systematic differences. For example, removing AI-side mentions could leave a bare retweet marker, and the affective transformation could introduce characteristic punctuation. We therefore implemented four diagnostics to test whether a domain-adapted readout exploited construction cues rather than provenance.

### 3.9.1 Surface-Feature Audit

We measured simple markers in the detection field of each condition: a bare retweet marker, the <USER> placeholder, any hashtag, the <URL> placeholder, an em dash, word count, and character count.

### 3.9.2 Artifact-Only Classifier

We fitted a logistic classifier on seven scalar features computed directly from the detection field: presence of a bare retweet marker matched by RT at a word boundary, presence of <USER>, presence of any hashtag, presence of <URL>, presence of an em dash, word count, and character count. The classifier used balanced class weights, was fitted on HEAD_TRAIN, and was evaluated on the dataset for both the full provenance task and the H0-versus-A0 contrast. Its purpose was to quantify how much apparent separation could be recovered from construction cues alone.

### 3.9.3 Artifact-Neutralisation Control

We repeated representation extraction, screening, fitting, and evaluation on a neutralised corpus. The transformation removed <URL> and <USER>, all hashtags, a leading or remaining bare retweet marker, and em or en dashes, then collapsed whitespace. Residual asymmetry was remeasured after neutralisation.

### 3.9.4 H0-versus-A0 Diagnostic Head

We fitted a second head, using the same modelling procedure as the primary head, on H0 and A0 rows from HEAD_TRAIN only. This contrast isolates original human text from factual machine-generated text. The diagnostic head was then applied to all four dataset conditions without refitting.

## 3.10 Statistical Analysis

All uncertainty estimates treat the semantic unit, not the individual text, as the resampling unit, preserving the dependence among the four conditions of a unit. The executed bootstrap draws $n$ semantic units with replacement from the $n$ units present, expands each drawn unit to all of its rows, and recomputes the statistic. 2,000 replicates are used with seed 42, and intervals are reported as the 2.5th and 97.5th percentiles of the replicate distribution. The same procedure is applied to single-model AUROC, to threshold-conditional recall and false-positive rate, and to paired differences between families.

For a comparison between two configurations scored on the same texts, the paired difference

$$\Delta\text{AUROC} = \text{AUROC}_A - \text{AUROC}_B$$

is recomputed within each bootstrap replicate using the same resampled units for both configurations, so that the interval reflects paired rather than independent variation. A difference is described as statistically distinguishable when the 95% interval excludes zero. For the paired transformation shifts we additionally report the Wilcoxon signed-rank test over unit-level differences, together with the mean shift and its bootstrap interval. No multiple-testing correction was applied; given the number of comparisons and the large sample, we therefore emphasise effect magnitude over significance.

## 3.11 Implementation

All computation ran on an NVIDIA H200 GPU with Python 3.12.9, PyTorch 2.6.0+cu126, and Transformers 4.56.2. Language models were loaded in bfloat16 without quantisation from a persistent on-cluster Hugging Face cache. Because Gemma-2-9B has a 256,000-token vocabulary, its detector batch size was reduced from 16 to 2; Llama used a batch size of 4. All reductions in both statistics were masked per sequence, so per-text scores were invariant to batch composition.

# 4. Results

## 4.1 Detector Performance at Native Operating Points

We first evaluated the released social-media detector on the dataset at its own decision rule. OSM-Det achieved AUROC 0.521 (95% CI 0.511 to 0.531) and AUPRC 0.530, close to the chance value of 0.5 for this balanced dataset. At its native argmax rule it flagged 39.0% of AI texts while also flagging 37.8% of human texts, giving accuracy 0.506 and balanced accuracy 0.506. The detector therefore does not separate the classes on this domain, and its default rule is additionally very liberal: it would raise false alarms on more than a third of genuine human disaster posts.

The only other configuration with a legitimate published threshold is Binoculars with the Falcon pair. Applied unchanged, it flagged 1.8% of AI texts at a 1.4% human false-positive rate, an extremely conservative operating point by contrast that detects almost nothing. In addition, the two judge configurations with a predeclared forced-choice rule behaved differently from one another: Llama flagged 13.3% of AI texts at a 19.2% false-positive rate, while Gemma flagged 1.6% at 1.6%.

### 4.2 Cross-Family Variation in Off-the-Shelf Detectors

Across the five families evaluated with Fast-DetectGPT, AUROC ranged from 0.421 (Qwen) to 0.517 (Gemma); with Binoculars it ranged from 0.417 (Qwen) to 0.492 (Falcon) (Table 5; Figure 1). Only one of these ten configurations exceeded 0.5, and then by only 0.017. Substituting the underlying language-model family therefore changed score behaviour without producing operationally useful separation.

Table 5. Threshold-independent performance of each detector configuration on the benchmark.

| Strategy | Family | AUROC | 95% CI | AUPRC | TPR @ 1% FPR | TPR @ 5% FPR |
|---|---|---|---|---|---|---|
| Fast-DetectGPT | Falcon | 0.492 | 0.483 to 0.501 | 0.500 | 0.011 | 0.052 |
| | Qwen | 0.421 | 0.412 to 0.430 | 0.445 | 0.007 | 0.033 |
| | Mistral | 0.435 | 0.426 to 0.445 | 0.457 | 0.011 | 0.040 |
| | Llama | 0.455 | 0.446 to 0.464 | 0.472 | 0.010 | 0.046 |
| | Gemma | 0.517 | 0.508 to 0.526 | 0.517 | 0.010 | 0.060 |
| Binoculars | Falcon | 0.492 | 0.483 to 0.501 | 0.506 | 0.015 | 0.059 |
| | Qwen | 0.417 | 0.408 to 0.426 | 0.447 | 0.010 | 0.036 |
| | Mistral | 0.429 | 0.419 to 0.438 | 0.454 | 0.011 | 0.042 |
| | Llama | 0.451 | 0.442 to 0.460 | 0.473 | 0.011 | 0.051 |
| | Gemma | 0.491 | 0.481 to 0.500 | 0.503 | 0.011 | 0.056 |
| LLM-as-Judge | Qwen | 0.402 | 0.393 to 0.410 | 0.440 | 0.006 | 0.034 |
| | Mistral | 0.413 | 0.405 to 0.421 | 0.437 | 0.005 | 0.023 |
| | Llama | 0.421 | 0.412 to 0.429 | 0.447 | 0.005 | 0.032 |
| | Gemma | 0.517 | 0.509 to 0.526 | 0.514 | 0.011 | 0.051 |
| OSM-Det baseline | - | 0.521 | 0.511 to 0.531 | 0.530 | 0.017 | 0.079 |
| OSM disaster linear head | - | 0.817 | 0.809 to 0.824 | 0.827 | 0.238 | 0.400 |

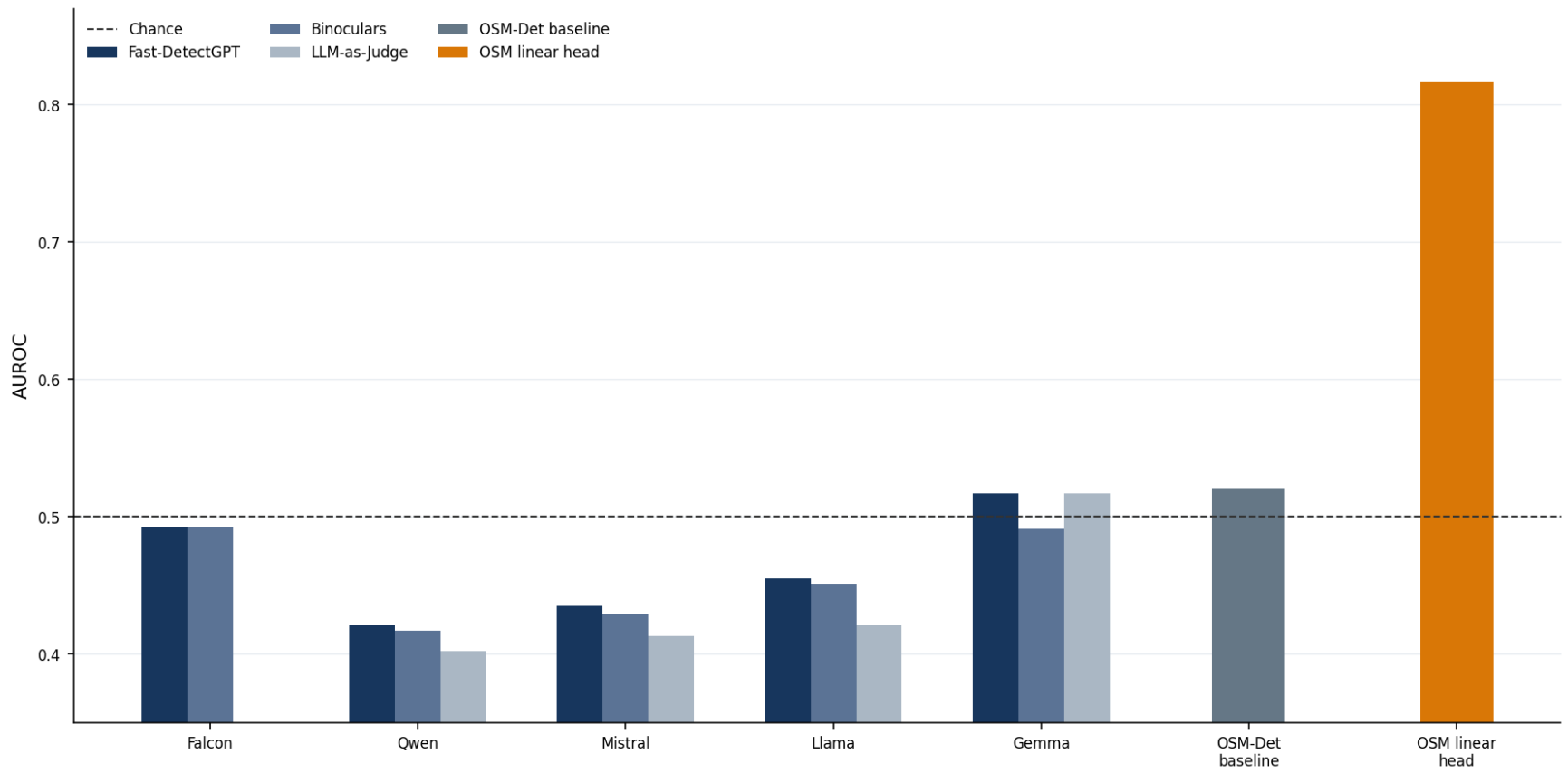

Figure 1. Benchmark AUROC by detection strategy and model family. The dashed line marks chance performance; the disaster-trained OSM linear head is highlighted.

Twelve of the fifteen frozen detector configurations fell below 0.5. Under the prespecified score orientation, those statistics ranked AI texts as more human-like than genuine human texts. Figure 2 shows that true-positive rates also remained low in the descriptive 5%-FPR region. Paired bootstrap comparisons confirmed that many family differences were statistically distinguishable (24 of 26 within-strategy comparisons had intervals excluding zero) but remained practically unhelpful. Under Fast-DetectGPT, for example, Gemma exceeded Qwen by 0.096 AUROC (95% CI 0.086–0.107), yet Gemma itself reached only 0.517.

Direct instruction-tuned authorship judgments did not outperform the likelihood-based statistics. Judge AUROC ranged from 0.402 (Qwen) to 0.517 (Gemma) (Table 5); Gemma was again the only judge above chance. Notably, Qwen was the weakest judge even though the AI dataset texts were generated with a Qwen backbone. Sharing a model family did not confer an advantage in this setting.

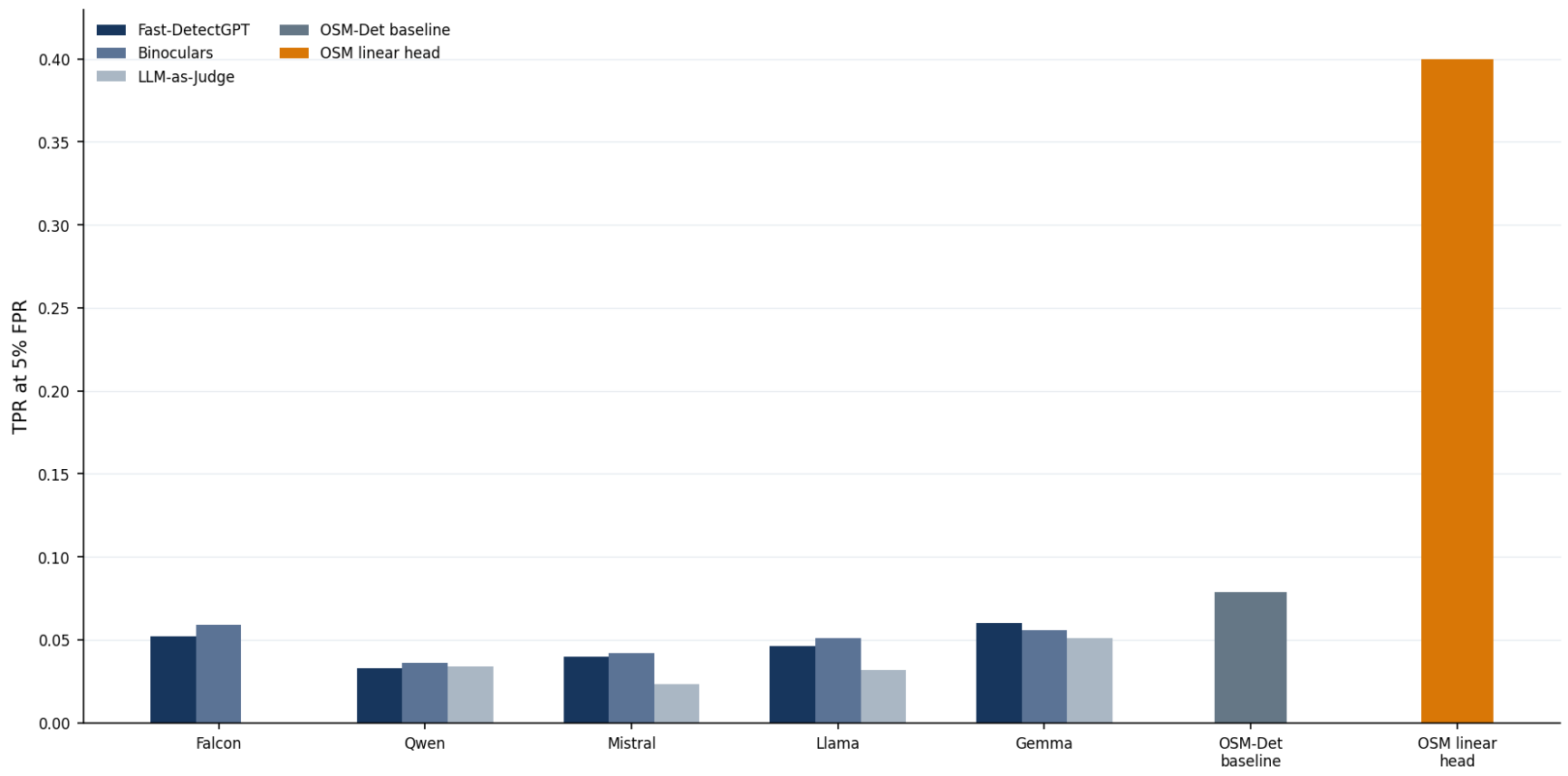

Figure 2. Descriptive true-positive rate at a 5% false-positive rate by detection strategy and model family.

## 4.3 Effect of Disaster-Domain Threshold Calibration

Disaster-domain calibration showed that weak frozen-detector performance could not be explained primarily by default thresholds. Calibration changes an operating point but not the underlying ordering, so AUROC and AUPRC remain unchanged. The prospective results in Table 6 show that sensitivity collapsed when false positives were constrained. Across the fourteen cross-family configurations, the highest recall at the calibration-derived $T_{FPR05}$ threshold was 3.6%, obtained by the Qwen LLM judge at a realised dataset FPR of 5.5%. Most other configurations detected only 1.5–3.3% of AI texts.

The balanced operating points demonstrate the opposite failure mode. Several configurations achieve apparently high recall only by classifying most human messages as AI-generated. Fast-DetectGPT with Qwen reaches 83.5% recall at an 88.3% FPR, Binoculars with Qwen reaches 87.5% recall at a 92.1% FPR, and LLM-as-Judge with Llama reaches 99.7% recall while incorrectly flagging 99.8% of human posts. Such operating points are unusable for disaster social sensing, where rejecting genuine eyewitness reports or requests for assistance may itself carry substantial cost. Thus, the primary limitation is not the absence of a favourable threshold, but that the human and AI score distributions overlap so strongly that increasing sensitivity necessarily produces unacceptable false alarms.

Figure 3 further shows how calibration exposes recall obtained through permissive native rules. OSM-Det detected 39.0% of AI texts at its native threshold but also flagged 37.8% of human texts. At the independently calibrated low-false-positive threshold, recall fell to 10.4% and the realised FPR to 6.7%.

Binoculars with Falcon remained highly conservative, detecting 1.8% of AI texts natively and 3.2% after calibration.

These results show that disaster-domain threshold calibration can control the location of the operating point, but it cannot rescue a detector whose scores contain little provenance information. In operational terms, the frozen detectors face an unfavorable choice: conservative thresholds protect genuine human reports but miss nearly all synthetic posts, whereas thresholds that recover high AI recall reject large fractions of authentic disaster communication. Text-only provenance detection therefore cannot be made suitable as a disaster-information gate merely by recalibrating its decision threshold.

Table 6. Native and disaster-calibrated operating performance. T_BAL, T_FPR05, and T_FPR01 were derived on the calibration corpus (N = 6,000 texts; 1,500 semantic units) with pair-balanced weights and applied unchanged to the dataset (N = 12,000 texts). Recall is the true-positive rate for the AI class and FPR is the false-positive rate for the human class. NA indicates that no legitimate native threshold existed for that configuration.

| **Strategy** | **Family** | **Native recall / FPR** | **T_FPR05** | **Test recall / FPR at $T_{FPR05}$** | **$T_{BAL}$ recall / FPR** | **$T_{FPR01}$ recall / FPR** |
|---|---|---|---|---|---|---|
| Fast-DetectGPT | Falcon | NA | 0.8599 | 3.3% / 2.9% | 4.2% / 3.8% | 0.5% / 0.3% |
| | Qwen | NA | 0.4913 | 2.1% / 3.3% | 83.5% / 88.3% | 0.4% / 0.4% |
| | Mistral | NA | 0.2114 | 1.5% / 1.6% | 0.6% / 0.6% | 0.4% / 0.3% |
| | Llama | NA | 0.6513 | 2.3% / 2.2% | 32.4% / 39.1% | 0.4% / 0.4% |
| | Gemma | NA | -0.8181 | 2.0% / 1.8% | 40.9% / 37.3% | 0.3% / 0.4% |
| Binoculars | Falcon | 1.8% / 1.4% | -0.8822 | 3.2% / 2.5% | 5.1% / 4.1% | 0.6% / 0.3% |
| | Qwen | NA | -0.9155 | 2.4% / 2.9% | 87.5% / 92.1% | 0.3% / 0.2% |
| | Mistral | NA | -0.9705 | 1.6% / 1.4% | 0.8% / 0.6% | 0.4% / 0.3% |
| | Llama | NA | -0.8929 | 2.3% / 2.1% | 2.9% / 2.6% | 0.3% / 0.2% |
| | Gemma | NA | -1.116 | 2.2% / 1.9% | 41.1% / 41.4% | 0.3% / 0.4% |
| LLM-as-Judge | Qwen | 5.5% / 8.5% | 0.7773 | 3.6% / 5.5% | NA / NA | 0.5% / 0.8% |
| | Mistral | 2.5% / 5.4% | 0.5 | 2.4% / 5.2% | NA / NA | 0.3% / 0.6% |
| | Llama | 13.3% / 19.2% | 0.9627 | 0.8% / 1.4% | 99.7% / 99.8% | 0.0% / 0.2% |
| | Gemma | 1.6% / 1.6% | 0.2942 | 2.4% / 2.4% | 1.2% / 1.1% | 0.2% / 0.1% |
| OSM-Det baseline | - | 39.0% / 37.8% | 0.9972 | 10.4% / 6.7% | 90.2% / 86.1% | 2.4% / 1.4% |
| OSM disaster linear head | - | NA | 0.8312 | 28.1% / 1.7% | 57.8% / 14.1% | 15.6% / 0.3% |

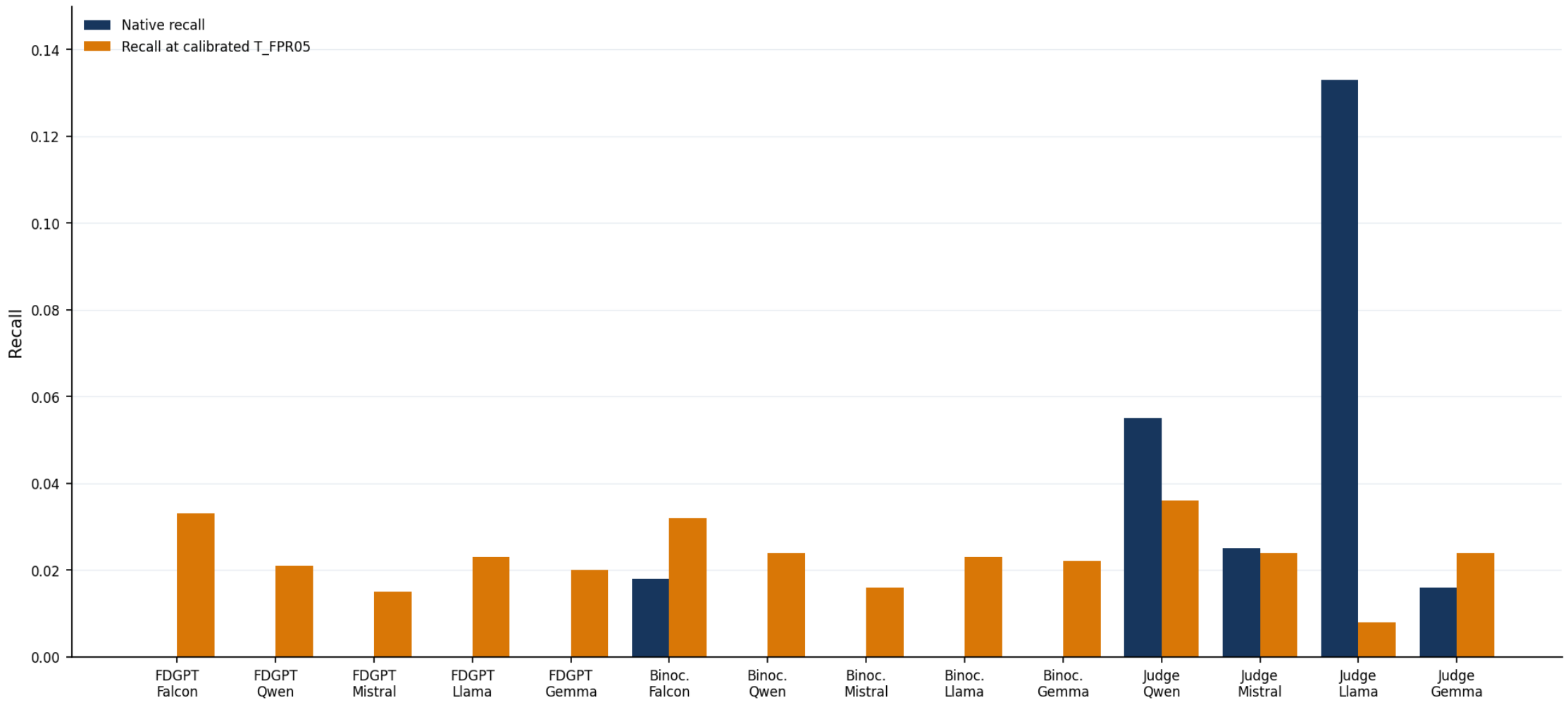

Figure 3. Recall at native operating points and at the independently calibrated $T_{FPR05}$ threshold. Native recall is plotted as zero when no legitimate native threshold exists.

## 4.4 Disaster-Calibrated OSM Linear Head

Supervised adaptation of the OSM-Det head produced a markedly different result from threshold calibration alone. The selected linear readout achieved AUROC 0.817 (95% CI 0.809–0.824) and AUPRC 0.827 on the event-disjoint dataset, compared with AUROC 0.521 for the off-the-shelf classifier (Table 5). At the independently calibrated $T_{FPR05}$ operating point, the adapted head detected 28.1% of AI texts at a realised human false-positive rate of 1.7% (Table 6). Supervised fitting therefore changed the ranking function and recovered substantially greater apparent separability.

The result was not produced by selecting a configuration on the final dataset. The linear head was fitted on HEAD_TRAIN, configuration selection was performed on event-disjoint HEAD_SELECTION, and threshold selection used HEAD_CALIBRATION; the dataset was used only for final evaluation. Among 312 combinations of hidden-state index, pooling strategy, regularisation strength, and standardisation, the selected configuration used hidden-state index 3, mean pooling, $C$=0.1, and no standardisation. Its AUROC was 0.819 on HEAD_SELECTION and 0.817 on the dataset, a difference of only 0.002. This close agreement shows that the apparent separation learned by the linear readout transfers strongly across the held-out event split rather than arising from direct test-set model selection.

Figure 4 shows that the result was not confined to one hidden layer. Mean-pooled representations remained strongly linearly separable across much of the frozen OSM-Det network, while classification-token separability rose after the embedding output and remained above chance. The selected intermediate representation therefore contained information that a linear classifier could exploit without changing encoder parameters.

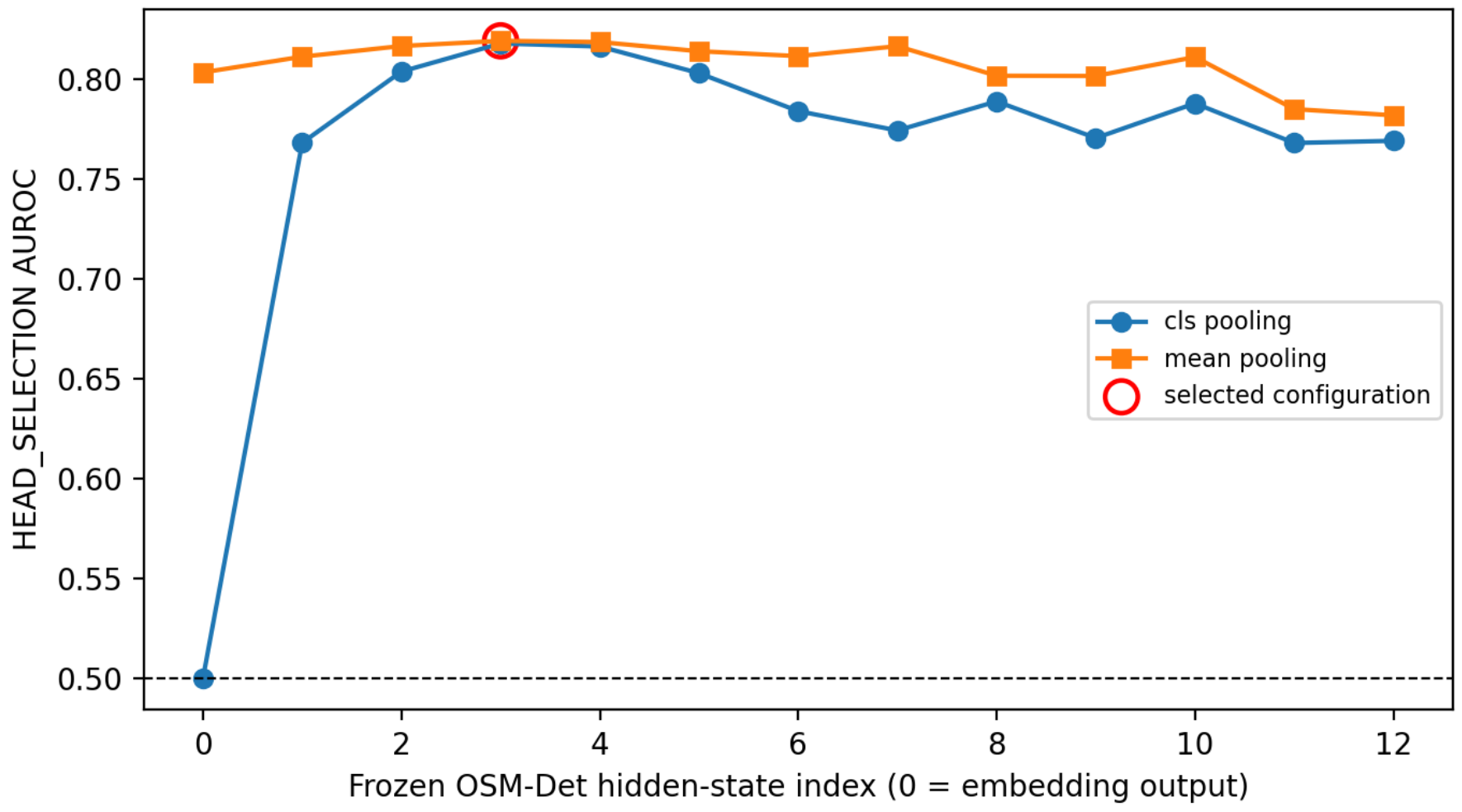

Figure 4. Layer-wise linear decodability of benchmark provenance labels from the frozen OSM-Det encoder. Each point is the best HEAD_SELECTION AUROC at that hidden-state index for a pooling strategy across the regularisation and standardisation grid. Index 0 is the embedding output; the circled point is the selected configuration.

### 4.5 Transformation Sensitivity

The paired design separates sensitivity to textual transformation from sensitivity to provenance. H0 and H1 share human provenance, and A0 and A1 share AI provenance; a provenance score robust to provenance-preserving edits should therefore change little within each pair. Table 7 and Figure 5 show substantial overlap among condition-specific score distributions, whereas Table 8 and Figure 6 show that paired changes vary in magnitude and direction by transformation and detector.

Table 7. Condition-specific behaviour of Gemma- and Llama-based configurations at the independently calibrated $T_{FPR05}$ threshold. For H0 and H1, the positive rate is an FPR; for A0 and A1, it is a TPR.

| Strategy | Family | Condition | Mean score | Median score | Positive rate | Interpretation |
|---|---|---|---|---|---|---|
| Binoculars | Gemma | A0 | -1.544 | -1.514 | 3.0% | TPR |
| | | A1 | -1.541 | -1.513 | 1.4% | TPR |
| | | H0 | -1.531 | -1.510 | 1.8% | FPR |
| | | H1 | -1.525 | -1.505 | 2.1% | FPR |
| Fast-DetectGPT | | A0 | -3.449 | -3.311 | 3.0% | TPR |
| | | A1 | -3.931 | -3.831 | 1.1% | TPR |
| | | H0 | -3.791 | -3.727 | 1.6% | FPR |
| | | H1 | -3.736 | -3.701 | 2.1% | FPR |
| LLM-as-Judge | | A0 | 0.037 | 0.004 | 3.2% | TPR |
| | | A1 | 0.020 | 0.002 | 1.6% | TPR |
| | | H0 | 0.028 | 0.002 | 2.5% | FPR |
| | | H1 | 0.026 | 0.002 | 2.3% | FPR |
| Binoculars | Llama | A0 | -1.245 | -1.234 | 2.9% | TPR |
| | | A1 | -1.214 | -1.202 | 1.7% | TPR |
| | | H0 | -1.206 | -1.188 | 1.8% | FPR |
| | | H1 | -1.193 | -1.178 | 2.5% | FPR |
| Fast-DetectGPT | | A0 | -1.403 | -1.411 | 2.8% | TPR |
| | | A1 | -1.459 | -1.429 | 1.9% | TPR |
| | | H0 | -1.307 | -1.276 | 1.9% | FPR |
| | | H1 | -1.234 | -1.233 | 2.5% | FPR |
| LLM-as-Judge | | A0 | 0.255 | 0.165 | 1.2% | TPR |
| | | A1 | 0.156 | 0.076 | 0.3% | TPR |
| | | H0 | 0.271 | 0.182 | 1.5% | FPR |
| | | H1 | 0.254 | 0.165 | 1.3% | FPR |

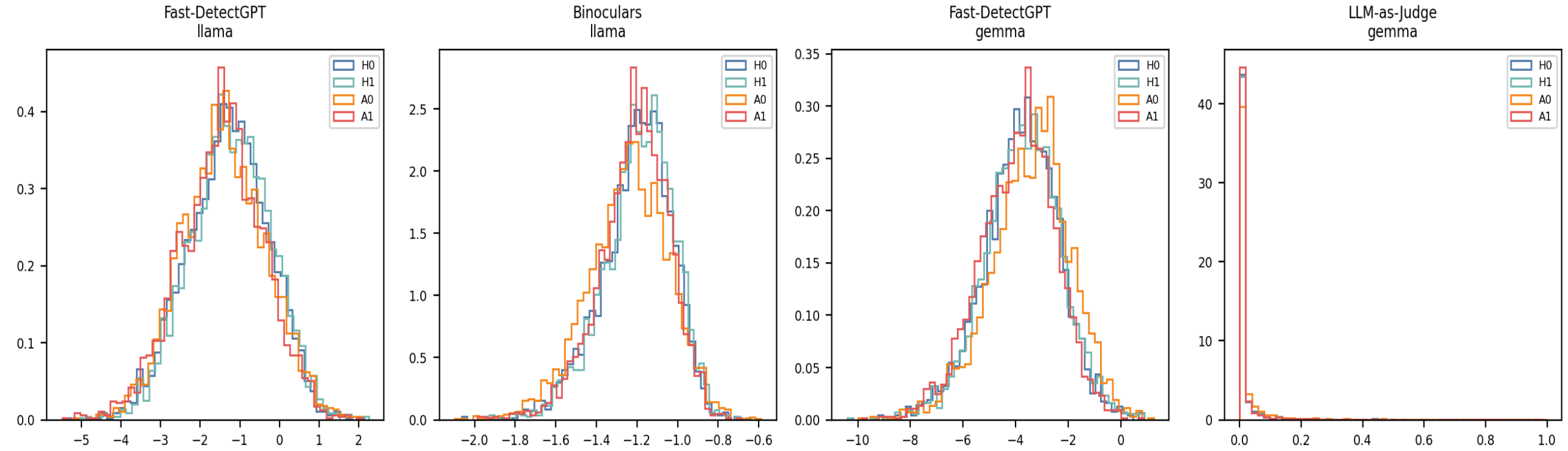

Figure 5. Condition-specific distributions of the AI-oriented score for selected Llama and Gemma configurations.

### 4.5.1 Minimal Human Proofreading

Minimal proofreading generally produced small shifts in detector scores. Among the Gemma and Llama configurations reported in Table 7, the largest mean H0-to-H1 change was 0.073 for Fast-DetectGPT with Llama (95% CI 0.060–0.086), while several configurations shifted by only a few thousandths or hundredths. Median changes are zero for almost all H0-to-H1 comparisons, consistent with both the constrained nature of the proofreading operation and the large proportion of H0/H1 pairs that are byte-identical.

Statistical significance is not equivalent to operational importance. Binoculars with Gemma, for example, had a mean shift of 0.006 (Wilcoxon $p = 3 \times 10^{-14}$), and the Llama LLM judge shifted by −0.017 ($p = 4 \times 10^{-27}$) (Table 8). With 3,000 paired units, small systematic changes are readily detectable even when their magnitude is limited. The disaster-trained OSM head achieved only AUROC 0.504 on H0 versus H1, indicating that minimal proofreading did not create a reliably separable class within genuinely human-authored posts.

Therefore, minimal LLM proofreading alone provides little evidence for reclassifying a genuinely human-origin message as synthetic. In other words, limited spelling, grammar, or punctuation correction does not change the underlying human provenance of a disaster report, and the detectors largely do not treat such editing as a strong provenance transition.

### 4.5.2 Controlled Affective Transformation

The A0-to-A1 transformation produced a qualitatively different pattern. Although provenance and factual content remained fixed, several scores moved substantially (Table 8). Fast-DetectGPT with Gemma showed the largest shift, decreasing by 0.482 on average (95% CI −0.515 to −0.448). The Llama LLM judge decreased by 0.100, whereas Binoculars with Llama moved in the opposite direction by 0.031. Binoculars with Gemma was effectively invariant: its mean shift was 0.004 and its confidence interval (−0.003 to 0.010) included zero. The scoring mechanisms therefore responded to different properties of the transformed text rather than to one invariant property of AI authorship.

The effect was also visible at the decision level. At $T_{FPR05}$, every Llama- and Gemma-based configuration in Table 7 detected a smaller fraction of A1 than A0. Fast-DetectGPT with Gemma fell from 3.0% on A0 to 1.1% on A1, the Gemma LLM judge from 3.2% to 1.6%, and the Llama LLM judge from 1.2% to 0.3%. A controlled change in framing could therefore determine whether an otherwise factually identical AI message crossed the operational boundary.

Thresholded changes did not always follow the mean score. Binoculars with Llama had a positive mean A0-to-A1 shift of 0.031, yet its detection rate fell from 2.9% to 1.7%. The transformation therefore reshaped the score distribution relative to the threshold rather than producing only a uniform location shift.

Table 8. Paired transformation effects for Gemma- and Llama-based configurations. Deltas are computed per semantic unit on the AI-oriented score; positive values indicate a more AI-like score after transformation.

| Strategy | Family | Contrast | Mean delta | 95% CI | Median | Wilcoxon p | Increased / decreased |
|---|---|---|---|---|---|---|---|
| Binoculars | Gemma | A0 to A1 | 0.004 | -0.003 to 0.010 | 0.009 | 0.319 | 1572 / 1428 |
| | | H0 to H1 | 0.006 | 0.004 to 0.009 | 0.000 | 3e-14 | 902 / 538 |
| Fast-DetectGPT | | A0 to A1 | -0.482 | -0.515 to -0.448 | -0.431 | 1e-133 | 939 / 2061 |
| | | H0 to H1 | 0.055 | 0.036 to 0.073 | 0.000 | 8e-18 | 910 / 530 |
| LLM-as-Judge | | A0 to A1 | -0.017 | -0.020 to -0.013 | -0.001 | 1e-131 | 803 / 2134 |
| | | H0 to H1 | -0.001 | -0.003 to -0.000 | 0.000 | 7e-11 | 483 / 800 |
| Binoculars | Llama | A0 to A1 | 0.031 | 0.026 to 0.035 | 0.030 | 2e-40 | 1856 / 1144 |
| | | H0 to H1 | 0.013 | 0.011 to 0.015 | 0.000 | 2e-30 | 912 / 528 |
| Fast-DetectGPT | | A0 to A1 | -0.056 | -0.081 to -0.031 | -0.017 | 0.001 | 1462 / 1538 |
| | | H0 to H1 | 0.073 | 0.060 to 0.086 | 0.000 | 1e-26 | 896 / 544 |
| LLM-as-Judge | | A0 to A1 | -0.100 | -0.105 to -0.093 | -0.052 | 3e-195 | 785 / 2208 |
| | | H0 to H1 | -0.017 | -0.020 to - | 0.000 | 4e-27 | 521 / 894 |
| | | | | 0.014 | | | |

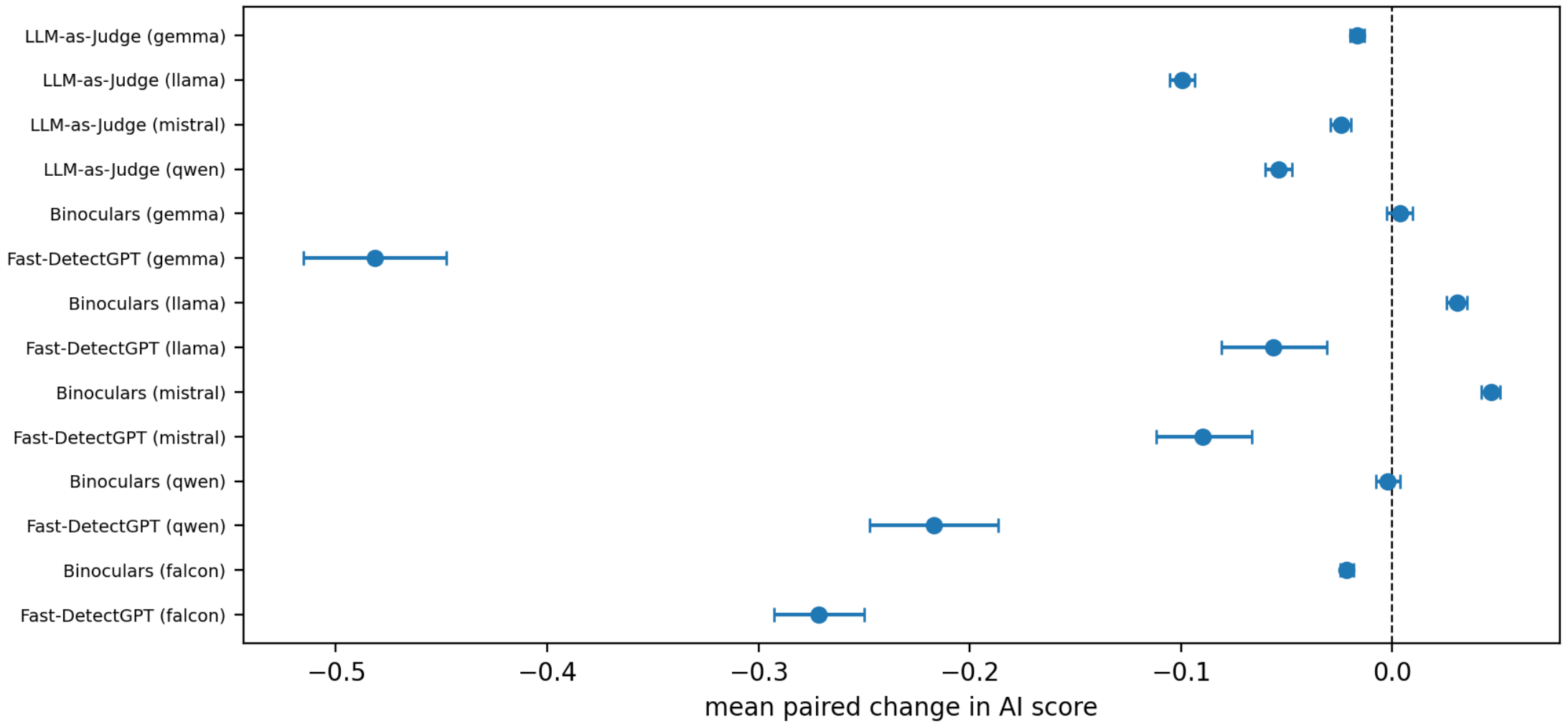

Figure 6. Mean paired change in the AI-oriented score under the controlled affective transformation from A0 to A1. Points show mean per-unit differences with 95% bootstrap intervals; zero indicates no change.

### 4.6 Artifact Diagnostics

The calibrated OSM linear head achieved AUROC 0.817, but this result does not by itself establish that the head detected authorship. The following analyses test whether superficial formatting and construction cues explain part of the apparent separation.

#### 4.6.1 Surface Asymmetries Between Conditions

A direct audit of the detection field identified systematic surface differences between the human and AI conditions that are consequences of corpus construction rather than of authorship (Table 9). The clearest is a bare retweet marker, present in 10.8% of A0 and A1 texts but only 0.0% of H0 and H1 texts, which arises because AI-only mentions were removed during cleaning and left the retweet prefix stranded. Mention placeholders appear in 42.8% of H0 texts but only 6.1% of A0 texts, and hashtags in 46.0% of H0 but 29.3% of A0. The em dash appears in 21.3% of A1 texts against 0.3% of A0 texts, a direct signature of the affective framing. Mean length also differs, with A0 the shortest condition and A1 the longest.

Table 9. Surface-marker prevalence by condition in the benchmark. Percentages are the share of texts containing each marker (N = 3,000 per condition).

| **Marker** | **H0** | **H1** | **A0** | **A1** |
|---|---|---|---|---|
| Bare retweet marker | 0.0% | 0.0% | 10.8% | 10.8% |
| <USER> placeholder | 42.8% | 43.0% | 6.1% | 6.1% |
| Any hashtag | 46.0% | 46.3% | 29.3% | 29.3% |
| <URL> placeholder | 79.0% | 79.0% | 79.0% | 79.0% |
| Em dash | 0.8% | 1.2% | 0.3% | 21.3% |
| Mean word count | 15.3 | 15.3 | 12.8 | 17.3 |
| Mean character count | 95.2 | 95.3 | 79.3 | 107.8 |

### 4.6.2 Artifact-Only Classification

To quantify how much apparent provenance separation could be recovered from superficial formatting cues alone, we fitted a logistic classifier using only seven surface features and no language representation. It achieved AUROC 0.784 on H0 versus A0 and 0.733 on the full four-condition task. On H0 versus A0, this exceeded the 0.733 achieved by the primary linear head and approached the 0.804 achieved by the diagnostic OSM head fitted specifically to that contrast. The largest fitted coefficient was attached to the bare retweet marker. Figure 7 compares these diagnostic results with the neutralised rerun.

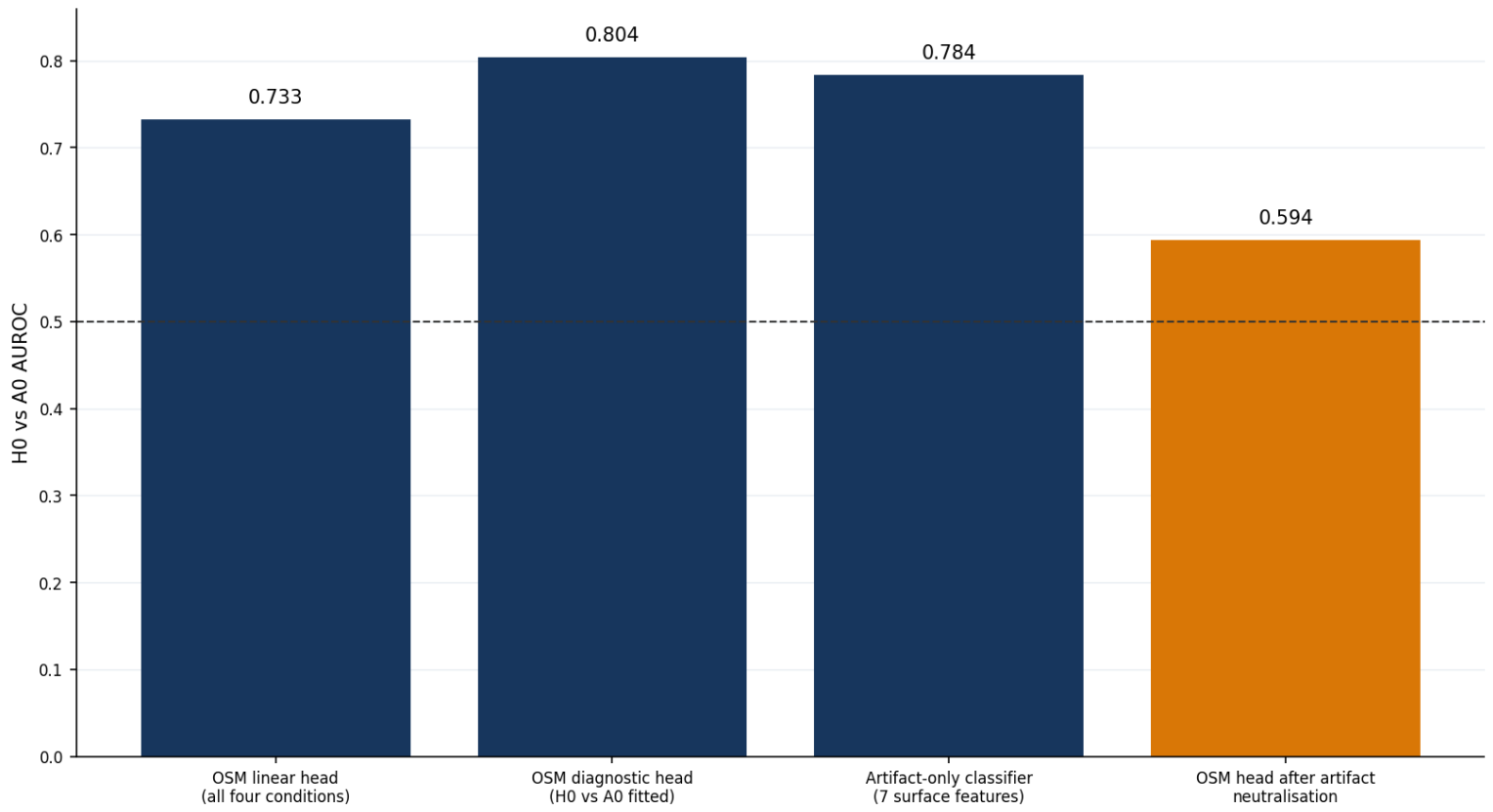

Figure 7. Artifact diagnostics for the factual provenance contrast (H0 versus A0). The dashed line marks chance performance; the orange bar shows the result after neutralisation.

### 4.6.3 Performance After Neutralising Identified Asymmetries

Removing the identified surface cues and rerunning the OSM pipeline reduced AUROC on H0 versus A0 from 0.733 to 0.594. Thus, 59.7% of the original above-chance separation disappeared after neutralisation. Overall four-condition AUROC fell from 0.817 to 0.735. In the neutralised corpus, bare retweet markers and mention placeholders were absent across conditions, and hashtag prevalence was below 0.2%.

## 4.7 Integrated Comparison Across Strategies and Model Families

Six observations summarise the results. First, the fourteen frozen cross-family configurations achieved AUROC 0.402–0.517, and only two exceeded chance; OSM-Det reached 0.521. Second, disaster-domain calibration relocated operating points without restoring sensitivity: the best prospective recall among the cross-family detectors at a controlled false-positive budget was 3.6%. Third, minimal proofreading produced small score shifts, whereas affective framing produced configuration-dependent changes despite unchanged provenance. Fourth, a supervised frozen-encoder readout reached AUROC 0.817. Fifth, a seven-feature classifier containing no neural representation matched or exceeded that readout on the H0-versus-A0 contrast, reaching 0.784 versus 0.733. Sixth, neutralising identified surface asymmetries reduced the supervised H0-versus-A0 result to 0.594.

## 5. Discussion

Current text-based AI detectors provided little reliable separation between human-authored and AI-generated disaster social-media messages. Across the fourteen frozen cross-family configurations, AUROC ranged from 0.402 to 0.517, while OSM-Det reached 0.521. Domain-specific calibration changed operating points but not ranking. At $T_{FPR05}$, the best cross-family configuration, the Qwen LLM judge, detected only 3.6% of AI texts at a realised 5.5% human false-positive rate. OSM-Det detected 10.4% at a realised 6.7% FPR, still missing almost nine in ten AI-generated messages.

Performance in the low-false-positive region is more informative for disaster social sensing than accuracy at a test-optimised threshold. Lowering a threshold cannot compensate for weak class separation: several balanced thresholds recovered high AI recall only by labelling most genuine human posts as AI. In a disaster stream, such false positives could suppress eyewitness observations or requests for assistance. Calibration can select a point on an error trade-off, but it cannot create provenance information that the score does not contain.

Generator matching did not produce a consistent advantage. A Qwen2.5-7B-Instruct backbone generated A0, yet Qwen was the weakest direct judge and performed poorly under both likelihood-based strategies. This result should not be interpreted as evidence that generator identity never matters; M4, RAID, DetectRL, MultiSocial, and social-media fine-tuning experiments all report substantial variation across generators, domains, platforms, and manipulations [28–33]. Rather, family identity alone was insufficient to overcome the joint shift to short, event-specific disaster posts. The non-transfer of some balanced thresholds across held-out events further shows that a rule tuned to one disaster distribution can collapse on another.

Threshold transfer was also incomplete across events. $T_{FPR05}$ was chosen to satisfy a 5% false-positive constraint on independent calibration events, but realised benchmark FPRs varied around that target. Because the event sets were disjoint, this difference reflects deployment-relevant shifts in entities, hazards, vocabulary, and posting conventions rather than test-set threshold tuning. The distinction between descriptive and prospective metrics is therefore important. For Fast-DetectGPT with Gemma, the benchmark ROC permits a descriptive TPR of 6.0% at exactly 5% FPR, whereas the independently selected threshold achieves 2.0% recall at a realised 1.8% FPR. The former describes what would be possible with benchmark labels; the latter estimates prospective operation.

The supervised head illustrates a central benchmark-validity risk. Its overall AUROC of 0.817 initially suggested useful domain adaptation, yet a logistic model using only seven surface features reached 0.784 on H0 versus A0, and neutralising identified cues reduced the head on that contrast from 0.733 to 0.594. Stylometric work confirms that controlled human and LLM samples can differ in lexical, grammatical, syntactic, and punctuation features [67], while the detector literature treats such patterns as an established signal family [42]. In the present dataset, the diagnostic evidence shows that much of the separation was tied to these features. Such cues may be helpful, but eventually they are not a dependable basis for operational provenance because their direction and prevalence can change after editing, platform processing, or a new generation workflow [62–64].

Minimal proofreading left many posts unchanged and produced only small score shifts, consistent with work that treats machine-polished text as distinct from fully machine-generated text [46]. Affective framing altered several scores much more strongly, although provenance and factual content remained fixed. Most reported Llama- and Gemma-based configurations detected a smaller fraction of A1 than A0, but Binoculars with Llama shifted in the opposite direction on the continuous score. Disaster-related emotional language is salient and can spread through social media [66]. Recent studies show that LLMs can perform strongly on structured emotional-intelligence tests and can generate responses that that are highly compassionate [68,69]. Our results are not sufficient to establish whether human authored posts are more emotional, or detectors recognize emotions in a universal way. Since the origins of symbolic language remain an open question, and that affective science continues to debate basic emotion and psychological-construction accounts [70,71], here we also choose to maintain a narrow interpretation.

Since text-based detection is insufficient, additional channels, such as account history, network coordination, geolocation, images, temporal consistency, platform-side generation records, watermarks, and signed credentials, may provide stronger provenance evidence, although watermarks require generation-time cooperation and can remain vulnerable to editing [55,56]. Future systems are encouraged to combine claim-level verification with source and network behaviour, temporal and spatial plausibility, and multimodal consistency. Future studies should also vary events, generators, prompts, temperatures, model sizes, languages, platforms, and post-editing intensity while recording human-AI edit histories to test whether converging evidence provides a more reliable basis for safeguarding disaster social sensing.

## 6. Conclusion

Generative AI does not eliminate the value of disaster social sensing, but it invalidates the assumption that writing style can establish who produced a post or whether its claims are reliable. On a paired, event-disjoint disaster benchmark, frozen text-only detectors remained near chance, and independent calibration yielded negligible recall at constrained false-positive operating points. A disaster-trained linear head appeared much stronger, yet artifact-only and neutralisation controls showed that much of its separation could be recovered from construction cues. Its sensitivity to two AI conditions with unchanged provenance further demonstrates that high dataset performance can reward transformation templates rather than authorship information.

Text-only AI-provenance detection should therefore not be deployed as a gatekeeper for disaster information. It may serve as one uncertain triage signal when thresholds are prospectively validated, uncertainty is visible, and downstream decisions can abstain. Trust should attach instead to corroborated claims, accountable sources, network behaviour, multimodal consistency, and machine-readable provenance when available. In the generative-AI era, disaster social sensing remains trustworthy only when systems verify evidence rather than infer trust from writing style alone.